\documentclass[final,3p,times]{elsarticle}

\usepackage{amssymb}
\usepackage{natbib}
\usepackage{amsmath}
\usepackage{amsthm}
\usepackage{xurl}
\usepackage{xcolor}
\usepackage{multirow}
\usepackage{hyperref}

\usepackage{caption}
\usepackage{subcaption}
\usepackage{graphicx}
\usepackage{tabularx}
\usepackage[section]{placeins}
\usepackage[shortcuts]{extdash}

\journal{Pattern Recognition}

\begin{document}

\begin{frontmatter}



\title{Conditional Motion In-betweening}

\author[inst1]{Jihoon Kim}
\ead{jihoon-kim@korea.ac.kr}
\author[inst1]{Taehyun Byun}
\ead{taehyun-byun@korea.ac.kr}
\author[inst2]{Seungyoun Shin}
\ead{2018112005@dgu.edu}
\author[inst3]{Jungdam Won}
\ead{jungdam@fb.com}
\author[inst1]{Sungjoon Choi\corref{cor1}}
\cortext[cor1]{Corresponding author}
\ead{sungjoon-choi@korea.ac.kr}


\affiliation[inst1]{organization={Department of Artificial Intelligence},
            addressline={Korea University}, 
            city={Seoul},
            postcode={02841}, 
            country={South Korea}}

\affiliation[inst2]{organization={Department of Computer Science and Engineering},
            addressline={Dongguk University}, 
            city={Seoul},
            postcode={04620}, 
            country={South Korea}}

\affiliation[inst3]{organization={Meta},
            addressline={1 Hacker Way}, 
            city={Menlo Park},
            postcode={94025}, 
            state={CA},
            country={United States}}

%
%
\begin{abstract}
Motion in-betweening (MIB) is a process of generating intermediate skeletal movement between the given start and target poses while preserving the naturalness of the motion, such as periodic footstep motion while walking. Although state-of-the-art MIB methods are capable of producing plausible motions given sparse key-poses, they often lack the controllability to generate motions satisfying the semantic contexts required in practical applications. We focus on the method that can handle pose or semantic conditioned MIB tasks using a unified model. We also present a motion augmentation method to improve the quality of pose-conditioned motion generation via defining a distribution over smooth trajectories. Our proposed method outperforms the existing state-of-the-art MIB method in pose prediction errors while providing additional controllability. Our code and results are available on our project web page: \url{https://jihoonerd.github.io/Conditional-Motion-In-Betweening}
\end{abstract}


\end{frontmatter}


\setcounter{figure}{0}

%
%
\section{Introduction}
\label{sec:introduction}

The demand for generating natural and expressive 3D human motion proliferates in the film and gaming industries. Despite the demand, however, generating diverse character movements in industries still dominantly relies on Motion Capture (MoCap) machines and traditional approaches \citep{zhao1994inverse, xiao2008automatic} rather than learning-based approaches due to its inherent complexity. A model that can synthesize natural human motion with providing controllability can help professional animators by allowing them to focus on creative and novel motions by reducing the labor involved in simple and redundant motion creation.
One unique property of human motion is that it is spatio-temporally constrained by a human's feasible skeletal (kinematic) structure, which distinguishes itself from static and structured data such as images. In the context of motion generation, a wide range of studies \citep{fragkiadaki2015recurrent, ghosh2017learning} has focused on forecasting natural motion-frames (i.e., motion prediction) from a given key-pose or multiple sets of key-poses. Others \citep{won2014generating, ahn2018text2action, guo2020action2motion} investigate methods to generate motion from given semantics such as textual action description. In this paper, we focus on a motion in-betweening (MIB) problem, where MIB is a process of generating skeletal motions that naturally interpolates a given set of key-poses, reducing the great amount of time and manual effort required for animators.

\begin{figure}[h]
  \centering
  \begin{subfigure}{0.475\textwidth}
    \centering
    \includegraphics[width=\textwidth]{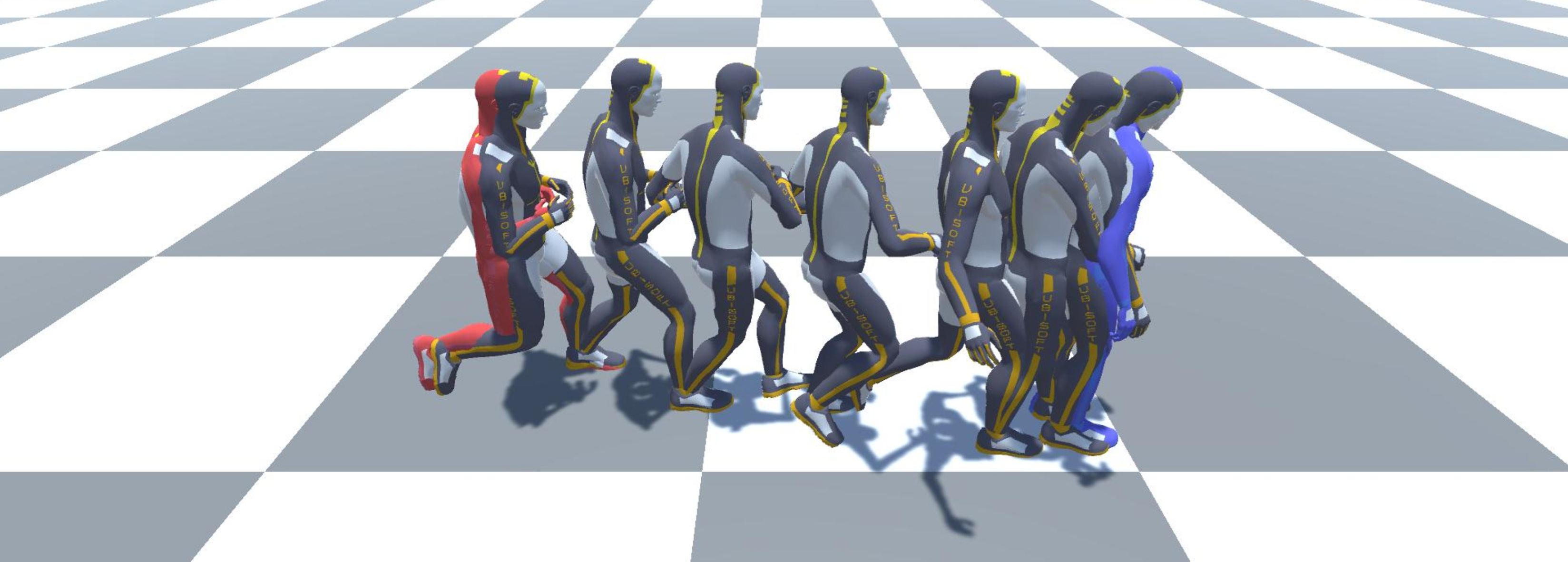}
    \caption{Ground truth}
  \end{subfigure}
  \begin{subfigure}{0.475\textwidth}
    \centering
    \includegraphics[width=\textwidth]{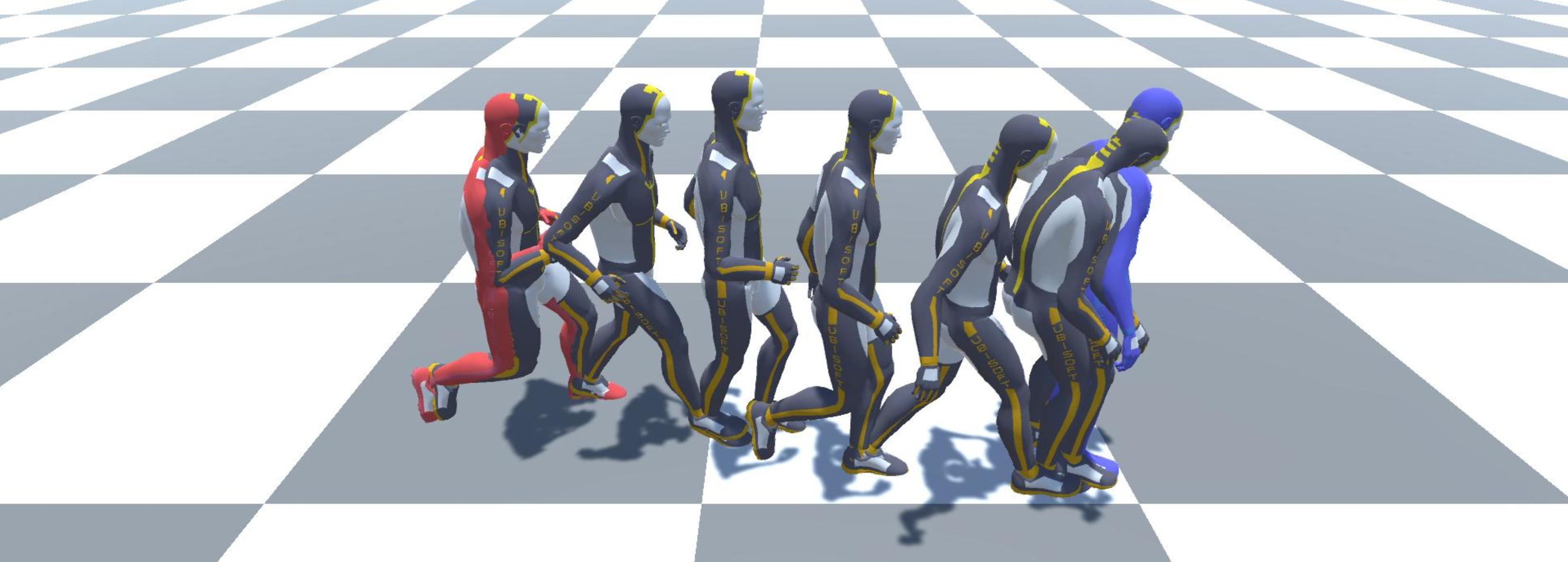}
    \caption{Motion in-betweening}
  \end{subfigure}
  \begin{subfigure}{0.475\textwidth}
    \centering
    \includegraphics[width=\textwidth]{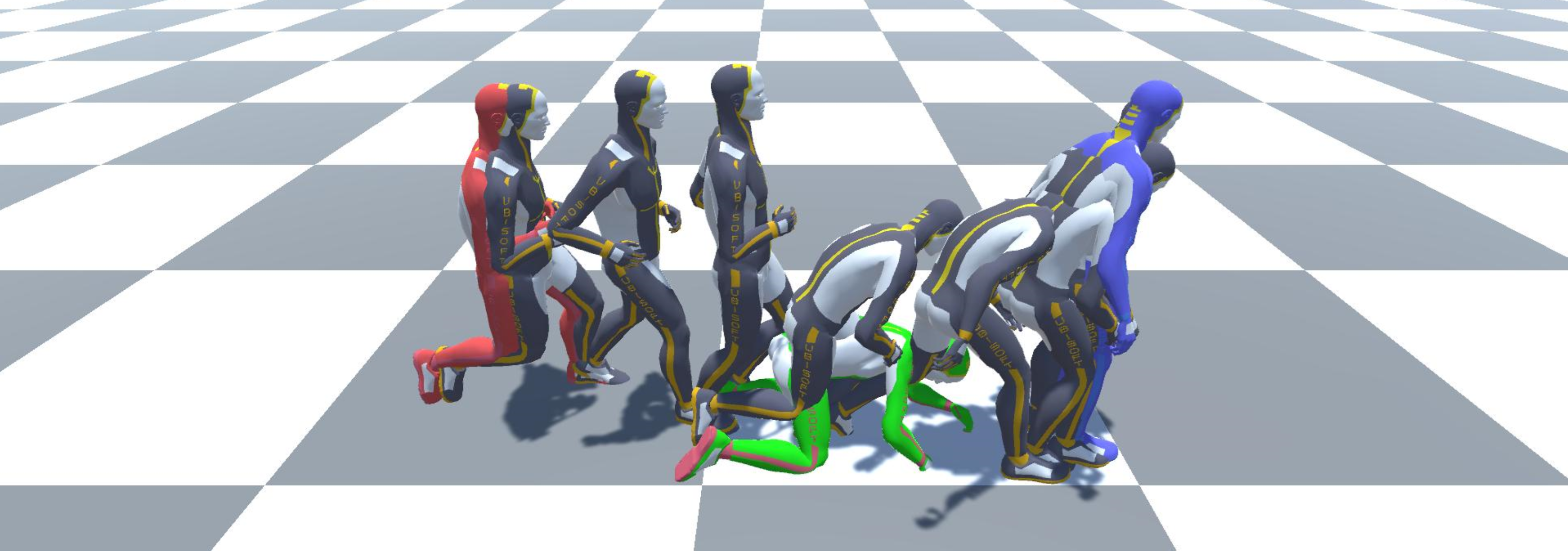}
    \caption{Pose-conditioned motion in-betweening}
  \end{subfigure}
  \begin{subfigure}{0.475\textwidth}
    \centering
    \includegraphics[width=\textwidth]{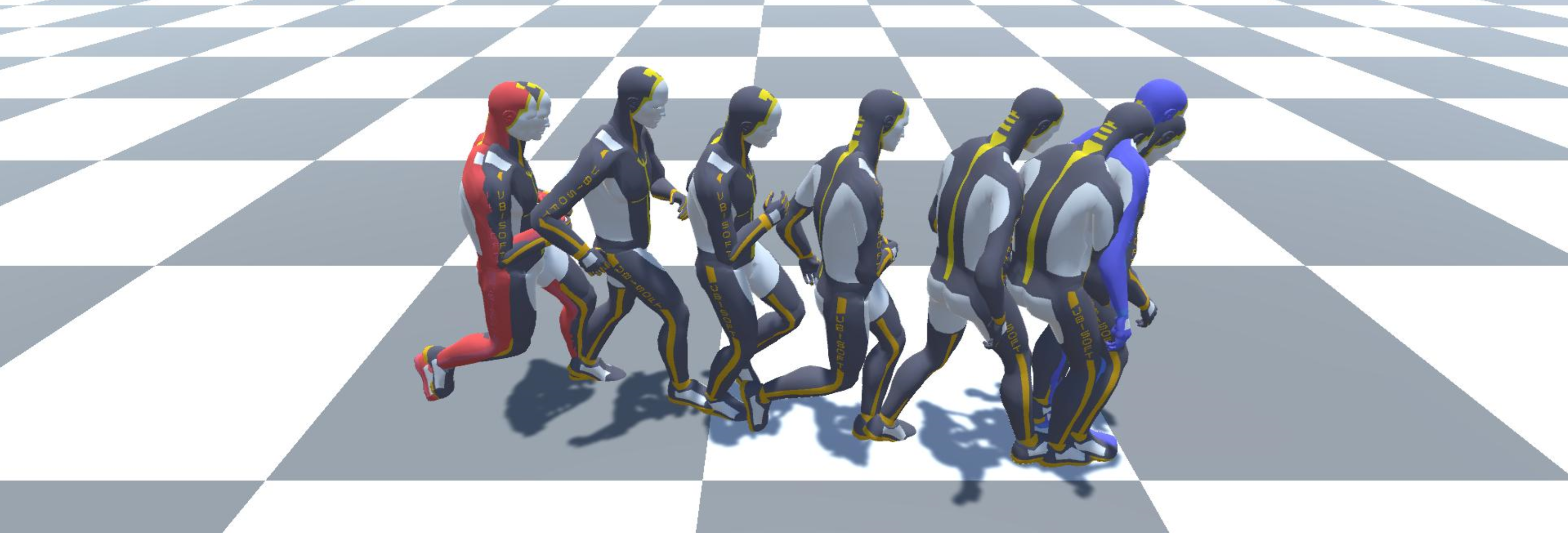}
    \caption{Semantic-conditioned motion in-betweening}
  \end{subfigure}
  \caption{(a) Ground Truth motion is visualized from human Motion Capture (MoCap) data. (b) Motion in-betweening is a task that connects given poses with plausible motion. (c) Pose-conditioned motion in-betweening connects given starting and target poses while satisfying the given interim pose. (d) Semantic-conditioned motion in-betweening connects given starting and target poses while satisfying given semantic information, such as walking, jumping, and dancing.}
  \label{fig:problem_description}
\end{figure}

Recent studies \citep{harvey2020robust, kaufmann2020convolutional} have shown that learning-based methods can provide plausible solutions for MIB given sparse key-poses. However, they often lack controllability in human motion generation. This paper presents a conditional motion in-betweening (CMIB) method that can handle two types of conditioned motion generation tasks: pose-conditioned and semantic\-/conditioned MIB. First, the pose-conditioned MIB generates a sequence of poses from the start, anchor, and target poses so that the generated motion naturally interpolates the given poses while preserving the naturalness of the motion.
We also present a motion augmentation method to further provide the pose-controllability beyond the given motion dataset. In particular, we augment the motion trajectories by sampling the root trajectories from a Gaussian random path distribution \cite{choi2016gaussian} which defines a probability distribution over smooth trajectories in reproducing kernel Hilbert space. 
Finally, semantic-conditioned MIB enables users to choose desired semantics at the inference stage. These conditioning problems are described in Figure \ref{fig:problem_description}.

We propose a CMIB method that integrates pose and semantic CMIB in a single Transformer encoder-based architecture. We interpret MIB as a masked motion modeling problem and introduce a \textit{randomized shuffled anchor pose} which makes our motion encoder perform pose-conditioned MIB. We also introduce a \textit{semantic embedding} token by prepending a sequence's semantic context to motion representation. To the best of our knowledge, we are the first to propose a controllable motion generation method in the MIB framework.

Our main contributions are summarized as follows:
\begin{itemize}
    \item We introduce a Transformer encoder-based CMIB model that can perform the following conditional motion generation tasks using a single model:
    \begin{itemize}
        \item Pose-conditioned: Generate a motion sequence that satisfies the given anchor pose while interpolating start and target poses
        \item Semantic-conditioned: Generate a motion sequence with given semantic information (e.g., walk, run, or dance)
    \end{itemize}
    \item We propose a motion data augmentation strategy that can generate various trajectories from existing motion for pose-conditioned MIB tasks.
    \item Our proposed method outperforms the existing state-of-the-art MIB method with additional capabilities of conditioned motion generation. Furthermore, we also present a performance evaluation measure to validate semantic-conditioned MIB tasks.
\end{itemize}

%
%
\section{Related Work}
\label{sec:related_work}

\textbf{Motion Prediction} Motion prediction refers to forecasting plausible motion from previous pose(s). The recent development of deep learning-based approaches has allowed significant advancement in motion synthesis. Recent studies adopted Recurrent Neural Networks (RNN) to address motion prediction problems. Fragkiadaki et al. \cite{fragkiadaki2015recurrent} introduced a three-layer long short-term memory (LSTM-3LR) network and encoder-recurrent-decoder (ERD) structure, which takes advantage of the autoregressive approach in latent dimension formulated from nonlinear encoder and decoder. Martinez et al. \cite{martinez2017human} suggested a sequence to sequence (seq2seq) architecture with residual connections in decoders. Ghosh et al. \cite{ghosh2017learning} integrated dropout autoencoder (DAE) to LSTM-3LR to mitigate the accumulation of error in RNN. Gui et al. \cite{gui2018adversarial} introduced a new loss based on geodesic loss instead of conventional Euclidean loss with adversarial training to predict human motion. Harvey et al. \cite{harvey2018recurrent} improved ERD by employing additional encoder blocks to inject various contexts to RNN. Aksan et al. \cite{aksan2019structured} decomposed motion prediction into joint-level prediction by modeling dependencies across joints with a structured prediction layer. Rossi et al. \cite{rossi2021human} have studied motion prediction from the perspective of trajectory prediction by leveraging both LSTM and generative adversarial network (GAN).

However, even with the help of the improved RNN structure, long-term prediction is an issue due to the inherent error accumulation problem of RNN, also with difficulty in parallelization. Aksan et al. \cite{aksan2020spatio} introduced self-attention to directly attend to the previous context and capture dependencies across poses and presented that the Transformer architecture can produce a long motion sequence with reduced error accumulation. Martinez et al. \cite{martinez2021pose} proposed a seq2seq Transformer encoder and decoder model with Graph Convolutional Networks (GCN) before and after the Transformer block.

The conditioned motion generation focuses on generating motion from given semantic information, such as music or action. Semantic information of human motion has been widely studied in action recognition tasks, but conditioned motion generation has a difference in that it should be able to generate desired motions for given semantics. Won et al. \cite{won2014generating} designed a generate-and-rank approach to choreograph human motion from high-level semantics. Ahn et al. \cite{ahn2018text2action} proposed a method to generate motion from text input. They trained an autoencoder between the language and action, and the language encoder is separately trained again in an adversarial manner with the attention mechanism. Henter et al. \cite{henter2020moglow} set the stage for using normalizing flows on probabilistic and controllable motion generation. Ling et al. \cite{ling2020character} proposed motion variational autoencoder (motion-VAE), which is based on conditional variational autoencoder (CVAE), to achieve controllable motion generation. Guo et al. \cite{guo2020action2motion} also studied action-conditioned motion generation, allowing the RNN-based architecture to learn action by providing action code with pose vector while training CVAE. Since MIB is also a part of motion generation, studies on motion prediction form an important foundation for our research.

\textbf{Motion in-betweening} In this work, we consider motion in-betweening (MIB) as a motion generation task, completing natural motion from sparse pose information. This task is analogous to image in-painting  \citep{pathak2016context}. However, MIB has an additional difficulty in that it has to synthesize spatio-temporal motion data. More closely, the video in-painting \citep{zeng2020learning, escher2021fast} shares similarities in that it should take both spatial and temporal aspects into consideration to connect sparse information naturally.

Frame-based video interpolation \citep{bao2019depth} also addresses interpolation problems, connecting sparse image frames with image sequences learned from videos. Image-level interpolation has the advantage that it can handle not only motion but also spatially adjacent information such as background or interacting objects. However, utilizing human motion for 3D applications is difficult with this approach since motion is entangled with other information in image space and lacks depth information, which is essential for 3D motion representation.

Rose et al. \cite{rose1998verbs} employed radial basis function (RBF) to interpolate parameterized motions. Mukai et al. \cite{mukai2005geostatistical} proposed a statistical prediction method which optimizes frame-level interpolating kernels for a given parametric space to perform MIB. Lehrmann et al. \cite{lehrmann2014efficient} have shown the possibility of Markov models in MIB tasks. 

MIB can also be viewed as a boundary value problem (BVP) for multi-joints. Two-point BVP studies the solution of a differential equation constrained to the provided start and target conditions. Li et al. \cite{li1991solving} approached motion planning problems as a trajectory optimization problem for given initial and final conditions. Xie et al. \cite{xie2015toward} combined optimal planning algorithm with a two-point BVP framework to solve optimal motion planning problem.

As deep neural network approaches are introduced, there have been great improvements in MIB tasks. Harvey et al. \cite{harvey2020robust} demonstrated studio-quality MIB methods based on their motion transition network \cite{harvey2018recurrent}. Also, they leveraged the least-squares generative adversarial network (LSGAN) \cite{mao2017least} to make generated motions to be more natural. However, these auto-regressive approaches are prone to long-horizon generation as the error accumulates over time and, thus, is difficult to parallelize. To overcome this, Kaufmann et al. \cite{kaufmann2020convolutional} employed a convolutional autoencoder by representing motion data in a matrix that can be interpreted as an image and showed that a non-autoregressive method could produce comparable results in MIB without degrading visual results. In contrast to prior work, we propose a controllable motion generation method on top of the MIB framework. The most relevant work of our approach is Harvey et al. \cite{harvey2020robust}, which, in contrast, only performs MIB tasks without controllability.

%
%
\section{Problem Formulation}
\label{sec:problem_formulation}

\subsection{Motion Data Representation}
We describe a human pose with joint positions and joint rotations. Joint positions are expressed on a real-world scale. There are multiple ways to represent rotations, such as Euler angles, axis-angle representation, and quaternions. We compare the model's performance on the three rotation expressions in Table \ref{table:rotation-comp-l2p}. In our experiment, the Euler angle representation is highly unstable during training. Empirically, the quaternion format outperforms other rotation expression formats in both performance and training stability, so we choose quaternion vectors as our rotation representation as previous work \citep{harvey2020robust} did.

\begin{table}[h!]
\centering
\begin{tabular}{c|c|c|c|c|c}
    Length (frames) & 30 & 40 & 60 & 80 & 120\\ \hline
    Euler & 2.17 & 2.93 & N/A & N/A & N/A\\ \hline
    Axis-angle & 1.38 & 1.51 & 2.15 & 2.89 & 4.54 \\ \hline
    Quaternion & \textbf{1.19} & \textbf{1.35} & \textbf{1.98} & \textbf{2.70} & \textbf{4.25} \\ \hline
\end{tabular}
\caption {Motion in-betweening L2P performance comparison by different rotation representations on the LAFAN1 dataset. ``N/A" indicates the setting was not able to train because of instability in the learning stage. Our CMIB model is used for evaluation.}
\label{table:rotation-comp-l2p}
\end{table}

We represent a joint position vector $\mathbf{p} \in \mathbb{R}^{3}$ in Euclidean space and adopt the quaternion $\mathbf{q} \in \mathbb{R}^{4}$ as a rotational representation of each joint. Positions and rotations at specific key-pose $t \in [1, T]$ for $J$ number of joints are denoted in a vectorized form $P_t \in \mathbb{R}^{3J}$ and $R_t \in \mathbb{R}^{4J}$, respectively. All positional and rotational values are represented in the global coordinate system in this paper.
 
\subsection{Conditional Motion In-betweening}
A wide range of studies have been conducted to generate motion, but motion in-betweening (MIB) has been less focused despite its practical usefulness in industries. MIB can be viewed as a boundary value problem for motion synthesis network $g$ that satisfies the initial conditions:

\begin{equation}
    g(t_1) = \begin{bmatrix} P_1, & R_1 \end{bmatrix} \quad
    g(t_T) = \begin{bmatrix} P_T, & R_T \end{bmatrix}
\end{equation}

Recent studies \citep{harvey2020robust, kaufmann2020convolutional} approached MIB as a problem conditioned only by both ends, but we extend the condition not only both ends but also additional intermediate frame and semantic information. In this work, we focus on two conditioning aspects: \textit{pose-conditioned} and \textit{semantic-conditioned} MIB. First, pose\-/conditioned motion in\-/betweening is a method of connecting starting and target pose while satisfying a given anchor pose as well. The other constraint is semantic information in motion. Motion data contains semantic information, such as emotions or action contents. Motion stylization \citep{holden2017fast, aberman2020unpaired} has covered methods to transfer style information while preserving motion's content features. Unlike motion stylization, our semantic-conditioned MIB needs to generate stylized content at once since it does not have available content motion frames other than starting and target pose.

%
%
\section{Proposed Method}
\label{sec:proposed_method}

\begin{figure}[htb]
    \centering
    \includegraphics[scale=0.8]{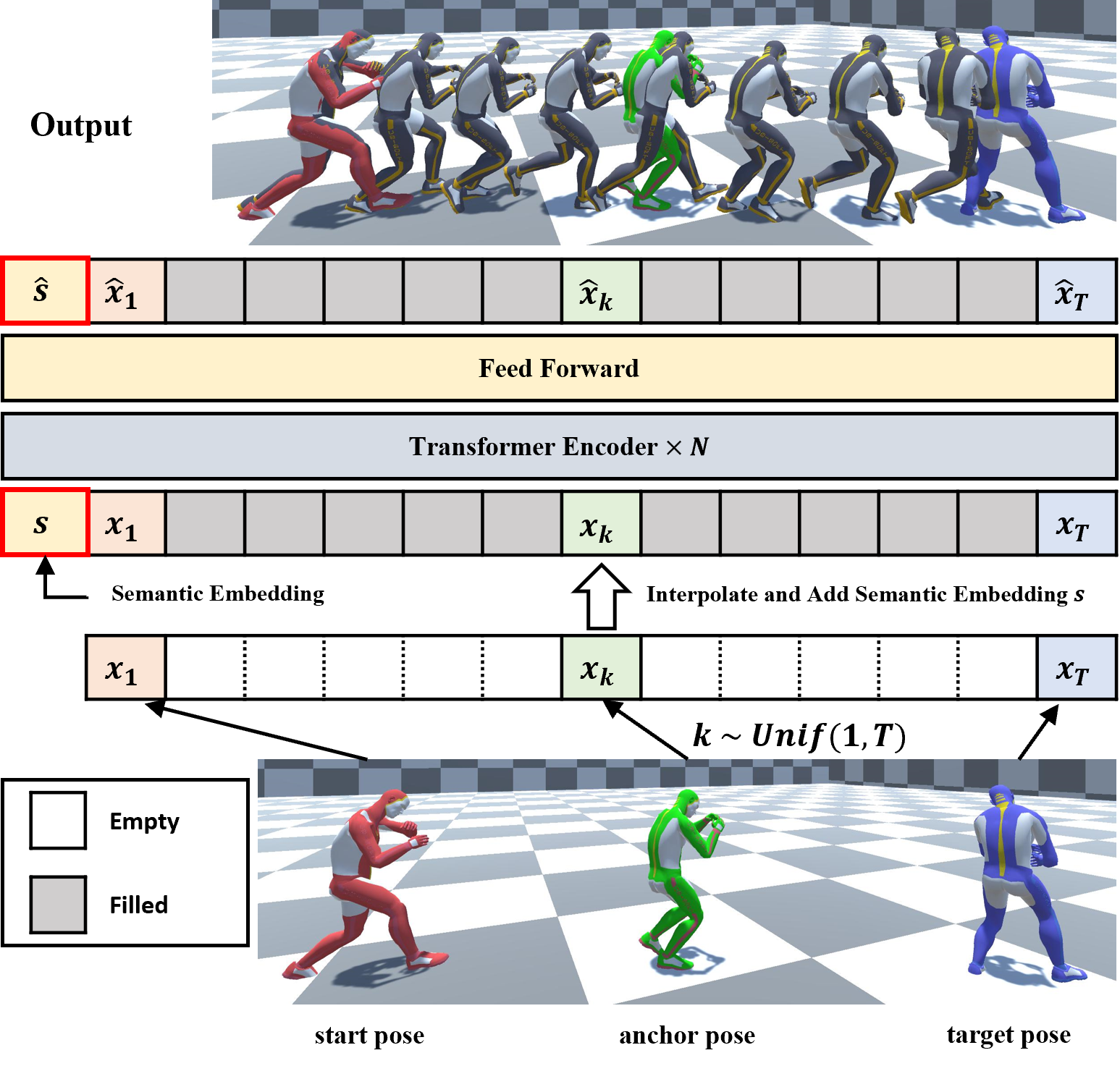}
    \caption{Overview of the model architecture. During training, we dynamically sample masking frames and replace those frames with interpolation from given poses. Then, the interpolated sequence is fed to the backbone Transformer encoder network with prepended semantic embedding. This task aims at generating natural human motion from given poses while providing semantic controllability.}
    \label{fig:model_overview}
\end{figure}

We propose a conditional motion in-betweening (CMIB) method that performs conditioned motion generation for both pose and semantic context, based on a Transformer architecture. Figure \ref{fig:model_overview} illustrates the overall architecture of the proposed method. 

%
%

%
%
\subsection{Pose Interpolation}

To perform MIB, starting, anchor (optional), and target poses should be provided. Instead of initializing key-poses with arbitrary values, we initialize our network with baseline interpolation to facilitate the learning process. For the given poses, the joint coordinates are interpolated with linear interpolation (Eq. \ref{eqn:lerp}), and joint rotations are interpolated by spherical linear interpolation (Eq. \ref{eqn:slerp}) in quaternion. Starting and target positions are denoted as $\mathbf{p}_{\mathrm{start}}, \mathbf{p}_{\mathrm{target}}$, and the position of the anchor pose given at $k$-th key-pose is written as $ \mathbf{p}_{k}$. Joint rotations of start ($\mathbf{q}_{\mathrm{start}}$), target ($\mathbf{q}_{\mathrm{start}}$), and anchor ($\mathbf{q}_{k}$) poses are represented in the same manner.

\begin{equation} \label{eqn:lerp}
    \operatorname{LERP}(\mathbf{p}_{\mathrm{start}}, \mathbf{p}_{\mathrm{target}}, \mathbf{p}_{k},t) = \begin{cases} \frac{k-t}{k-1} \mathbf{p}_{\mathrm{start}} + \frac{t-1}{k-1} \mathbf{p}_{k} & \text{for } 1 \leq t < k \\ \frac{T-t}{T-k} \mathbf{p}_{k} + \frac{t-k}{T-k} \mathbf{p}_{\mathrm{target}} & \text{for } k \leq t \leq T \end{cases}
\end{equation}
\begin{equation} \label{eqn:slerp}
    \operatorname{SLERP}(\mathbf{q}_{\mathrm{start}}, \mathbf{q}_{\mathrm{target}}, \mathbf{q}_{k}, t) = \begin{cases}  \frac{\sin \left( \frac{k-t}{k-1} \theta_1 \right)}{\sin \theta_1} \mathbf{q}_{\mathrm{start}} + \frac{\sin \left(  \frac{t-1}{k-1} \theta_1 \right)}{\sin \theta_1} \mathbf{q}_{k} & \text{for } 1 \leq t < k \\  \frac{\sin \left( \frac{T-t}{T-k} \theta_2 \right)}{\sin \theta_2} \mathbf{q}_{k} + \frac{\sin \left( \frac{t-k}{T-k} \theta_2 \right)}{\sin \theta_2} \mathbf{q}_{\mathrm{target}} & \text{for } k \leq t \leq T \end{cases}
\end{equation}

where $\theta_1 = \arccos (\mathbf{q}_{\mathrm{start}} \cdot \mathbf{q}_{k})$ and $\theta_2 = \arccos (\mathbf{q}_{k} \cdot \mathbf{q}_{\mathrm{target}})$.

%
%
\subsection{Model Architecture}

We utilize a Transformer encoder to implement CMIB. Since our objective is to create natural motion between given key-poses, we make the Transformer's multi-head self-attention to attend to all given input poses without masking. Input motion data is interpolated with LERP and SLERP before feeding into the motion encoder. Input motion can be compactly represented by vectorized positions $P_t \in \mathbb{R}^{3J}$ and rotations $R_t \in \mathbb{R}^{4J}$ at time step $t$:

\begin{equation}
    X = \begin{bmatrix} P_1, & R_1 \\ P_2, & R_2 \\ P_3, & R_3 \\\vdots & \vdots \\ P_{T}, & R_{T} \end{bmatrix} = \begin{bmatrix} \mathbf{x}_{1} \\ \mathbf{x}_{2} \\  \mathbf{x}_{3} \\ \vdots \\ \mathbf{x}_{T} \end{bmatrix}  \in \mathbb{R}^{T \times d} 
\end{equation}
where $d=J \cdot (3+4)$.

Since the non-autoregressive Transformer is leveraged, additional temporal information of the motion needs to be incorporated. In this work, we adopt learned positional embedding \cite{gehring2017convolutional}. The dimension of positional embedding is set to be equal to that of pose representation $\mathbf{x}_{t}$. Learned positional embedding is added before feeding into the Transformer. With this, we can write the input motion matrix $I$:
\begin{equation}
    I = \begin{bmatrix} \mathbf{x}_{1} + PE_{1}\\ \mathbf{x}_{2} + PE_{2} \\ \mathbf{x}_{3} + PE_{3} \\ \vdots \\ \mathbf{x}_{T} + PE_{T} \end{bmatrix}  \in \mathbb{R}^{T \times d}
\end{equation}
where $PE_{t} \in \mathbb{R}^{d}, 1 \leq t \leq T$.

Our motion encoder uses multiple encoder layers consisting of multi-head attention and position-wise feed-forward networks. Each single head attention is computed as:
\begin{align}
    Q &= W_{i}^{Q}I \\
    K &= W_{i}^{K}I \\
    V &= W_{i}^{V}I \\
    H_{i} &= \operatorname{softmax} \left( \frac{Q K^{\top}}{\sqrt{d_k}} \right) V
\end{align}
where $m$ is the number of heads, $d_k = {\frac{d}{m}}$ is the dimension of motion representation and $W_{i}^{Q}, W_{i}^{K}$ and $W_{i}^{V}$ are trainable parameter matrices for $i$-th head $H_{i}$. This single-head computation can be extended to the multi-head version by concatenating multiple single-heads followed by additional projection with $W^{O}$:
\begin{equation}
    \operatorname{MultiHead}(H_1, \cdots, H_m) = W^{O} \operatorname{Concat}(H_1, \cdots, H_m)
\end{equation}
where $W^{O} \in \mathbb{R}^{md \times d}$.
A fully connected feed-forward network (FCN) projects the encoder's representation dimension to the motion representation dimension. FCN block is composed of two stacked linear transformations with GeLU activation \cite{hendrycks2016gaussian}. Residual connection \cite{he2016deep} is used for both multi-head attention and FCNs followed by layer normalization. The Transformer encoder outputs a matrix of the same size as $I$, representing the complete predicted motion sequence.

Since our model learns positions and rotations in the global coordinate system, link lengths from the predicted output are not guaranteed to have the same lengths as the given kinematic chain. To make the link lengths consistent, we linearly scale the predicted link lengths to have the same lengths as the predefined kinematic chain.

%
%
\subsection{Training Phase}

%
%
\subsubsection{Randomized Shuffled Anchor Pose}

Pose conditioning is a way of generating motion to make it satisfy a given anchor pose during the MIB process. An anchor pose can be located between the start and target poses and it should be physically feasible in the context of MIB. This makes a clear difference from existing MIB methods, which perform MIB only for starting and target poses. Inspired from dynamic masking in RoBERTa \cite{liu2019roberta}, we uniformly sample the anchor key-pose from training motion sequences for every iteration. The sampled anchor pose is provided to the Transformer encoder along with the start and target poses after pose interpolation.

%
%
\subsubsection{Motion Data Augmentation}
\label{subsec:Data_Aug}

The quality of pose-conditioned motion in-betweening is affected by the motion path distribution of the training dataset. However, training motion data are often insufficient to fully cover various human motions. In the image processing domain, a number of data augmentation methods (e.g., random cropping or warping) are widely used, however, most of the existing augmentation methods in vision domains are not directly applicable to the motion domain due to their temporal characteristics. Harvey et al. \cite{harvey2018recurrent} used simple augmentation strategies of mirroring the motion data in the forward direction of the character. However, this method has a clear limitation in that it is not able to create a qualitatively different motion. Here, we introduce a motion augmentation method by first defining a probability distribution of smooth trajectories interpolating the start and target points. This method generates multiple motion trajectories from existing motions utilizing Gaussian Random Path (GRP) \cite{choi2016gaussian}.

Suppose that $(x_t,y_t) \in \mathbb{R}^2$ be a two-dimensional root joint position projected onto the ground plane extracted from given three-dimensional position vector $p_t\in \mathbb{R}^3$ and  $(\mathbf{x}_{\mathcal{P}},\mathbf{y}_{\mathcal{P}})=\left\{(x_t,y_t)\,|\,t=1,\dots,T\right\}$ be a two-dimensional trajectory of length $T$. 
Let $\left\{\mathbf{x}_a, \mathbf{y}_a\right\} = \left\{(x_1, y_1), (x_T,y_T)\right\}$ be a pair of start and target anchoring root positions. 
Our proposed augmentation method utilizes $\mathbf{x}_{\mathcal{P}}$ and anchoring root vectors $\left\{\mathbf{x}_a, \mathbf{y}_a\right\}$ to augment new root path $\tilde{\textbf{y}}_{\mathcal{P}}$.
Note that we assume the monotonicity of $\mathbf{x}_{\mathcal{P}}$.
Given a squared exponential kernel function $\mathbf{k}(\mathbf{x},\mathbf{x}') = \exp(-(\mathbf{x}-\mathbf{x}')^2/{2})$ and $\mathbf{x}_{\mathcal{P}}$, a Gaussian random path distribution $\tilde{\mathbf{y}}_{\mathcal{P}}$ is specified with a mean path $\mu_{\mathcal{P}}$ and a covariance matrix $K_{\mathcal{P}}$ as following:
\begin{equation} \label{eqn:10}
    \tilde{\textbf{y}}_{\mathcal{P}}\sim\mathcal{N}(\mu_{\mathcal{P}}, K_{\mathcal{P}}), 
\end{equation}
where 
\begin{align}
    \mu_{\mathcal{P}} &= \textbf{k}(\textbf{x}_{\mathcal{P}}, \textbf{x}_a)(\textbf{K}_a + \sigma^2_wI)^{-1}\textbf{y}_a, \\
    K_\mathcal{P} &= \textbf{K}_{\mathcal{P}} - \textbf{k}(\textbf{x}_{\mathcal{P}}, \textbf{x}_a) (\textbf{K}_a + \sigma^2_wI)^{-1}\textbf{k}(\textbf{x}_{\mathcal{P}}, \textbf{x}_a)^T, 
\end{align}
$\mathbf{k}(\mathbf{x}_{\mathcal{P}}, \mathbf{x}_a) \in \mathbb{R}^{T\times2}$ is a kernel matrix of $\mathbf{x}$'s path indices and anchored start and target. The kernel matrix of start and target indices is represented as $\mathbf{K}_a = \mathbf{K}(\mathbf{x}_a,\mathbf{x}_a) \in \mathbb{R}^{2\times2}$, and $\mathbf{K}_{\mathcal{P}} =\mathbf{K}(\mathbf{x}_{\mathcal{P}}, \mathbf{x}_{\mathcal{P}}) \in \mathbb{R}^{T\times T}$ indicates a kernel matrix of given $\mathbf{x}_{\textrm{path}}$. In the sense of Cholesky decomposition of $\mathbf{LL}^T=K_\mathcal{P}$, Eq. \ref{eqn:10} is equivalent to $\tilde{\mathbf{y}}_\mathcal{P}=\mu_{\mathcal{P}}+\mathbf{Lu}$, where $\mathbf{u}\sim \mathcal{N}(0,I)$. 

\begin{figure}[h]
  \centering
  \begin{subfigure}{0.475\textwidth}
    \centering
    \includegraphics[width=\textwidth]{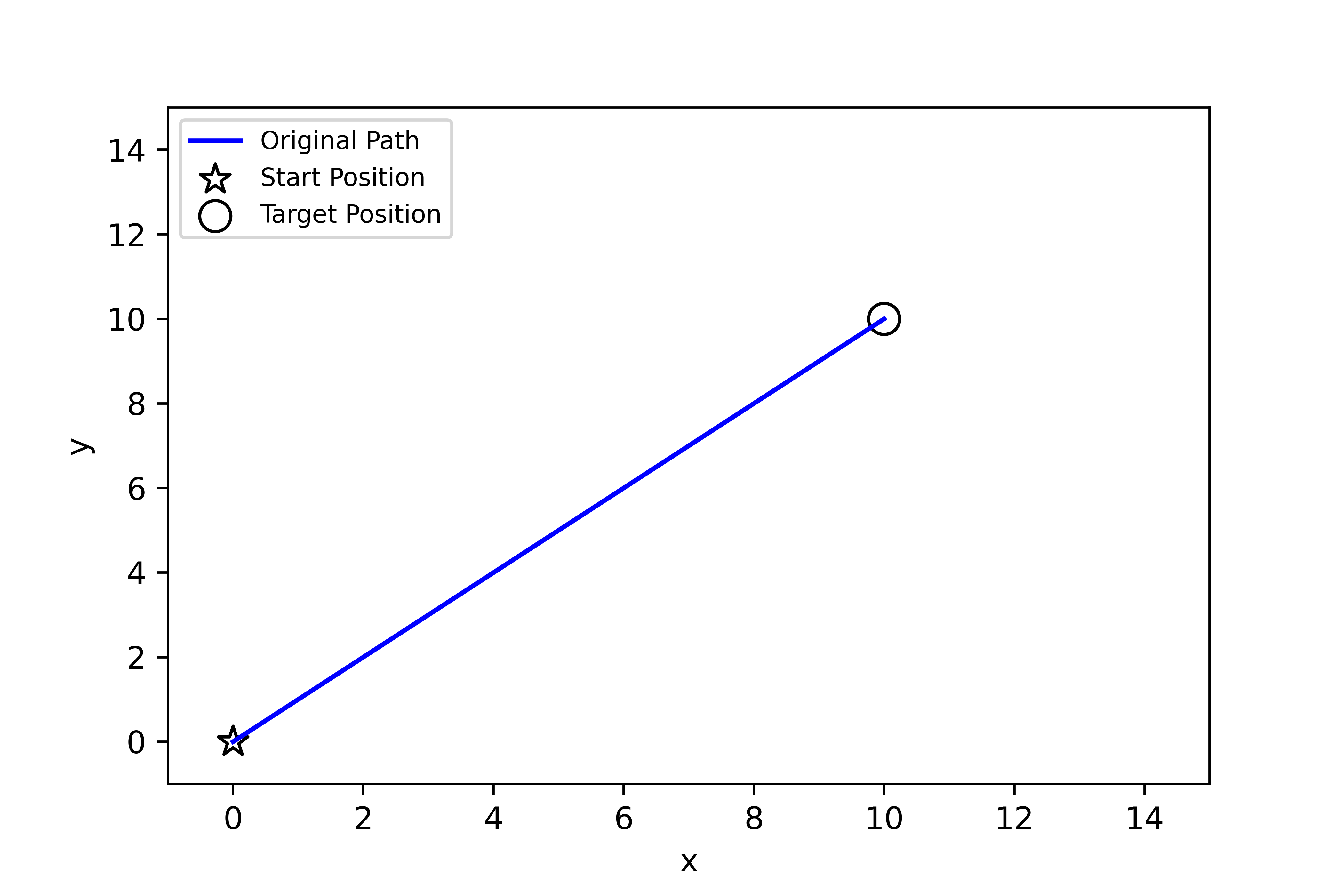}
    \caption{}
  \end{subfigure}
  \begin{subfigure}{0.475\textwidth}
    \centering
    \includegraphics[width=\textwidth]{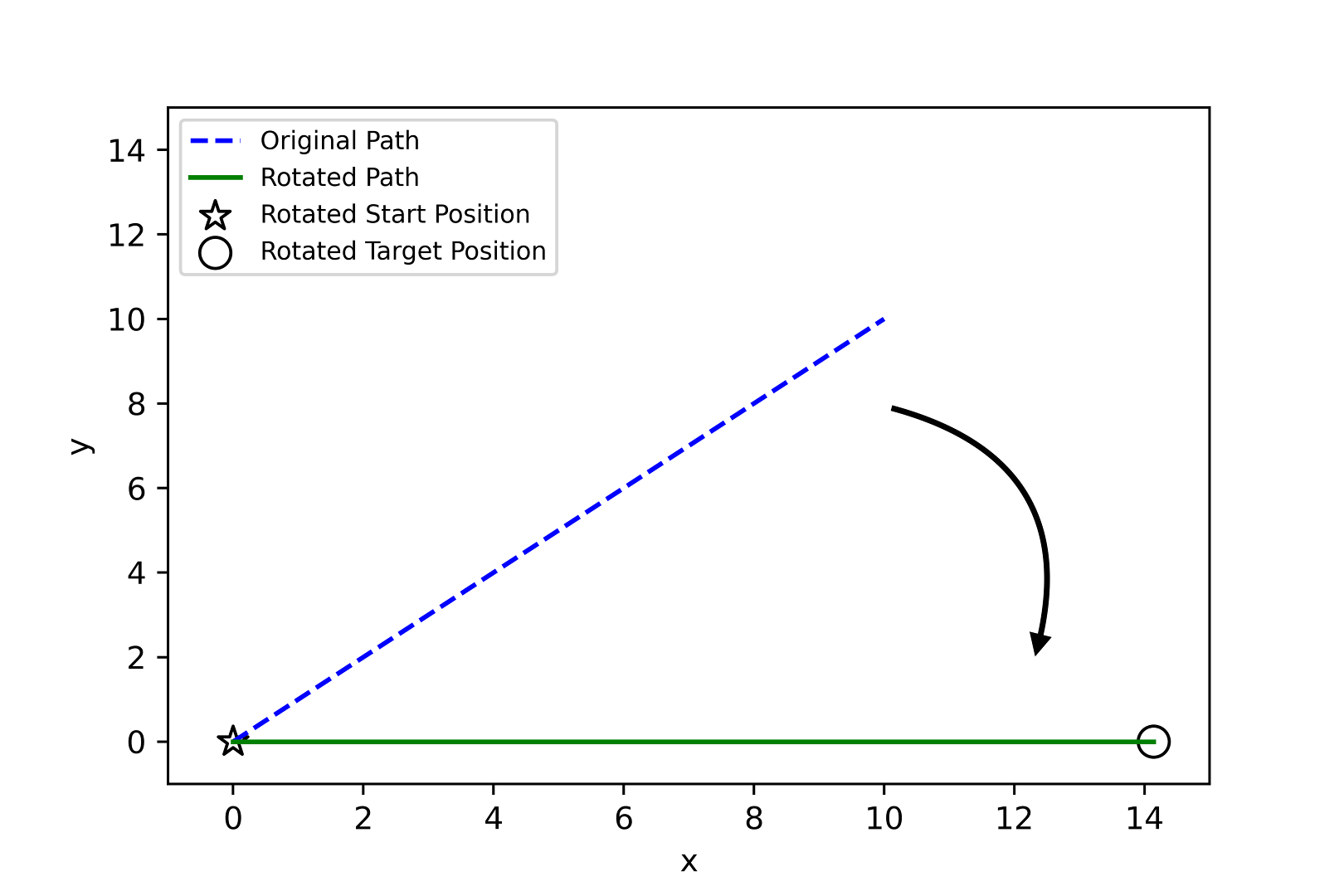}
    \caption{}
  \end{subfigure}
  \begin{subfigure}{0.475\textwidth}
    \centering
    \includegraphics[width=\textwidth]{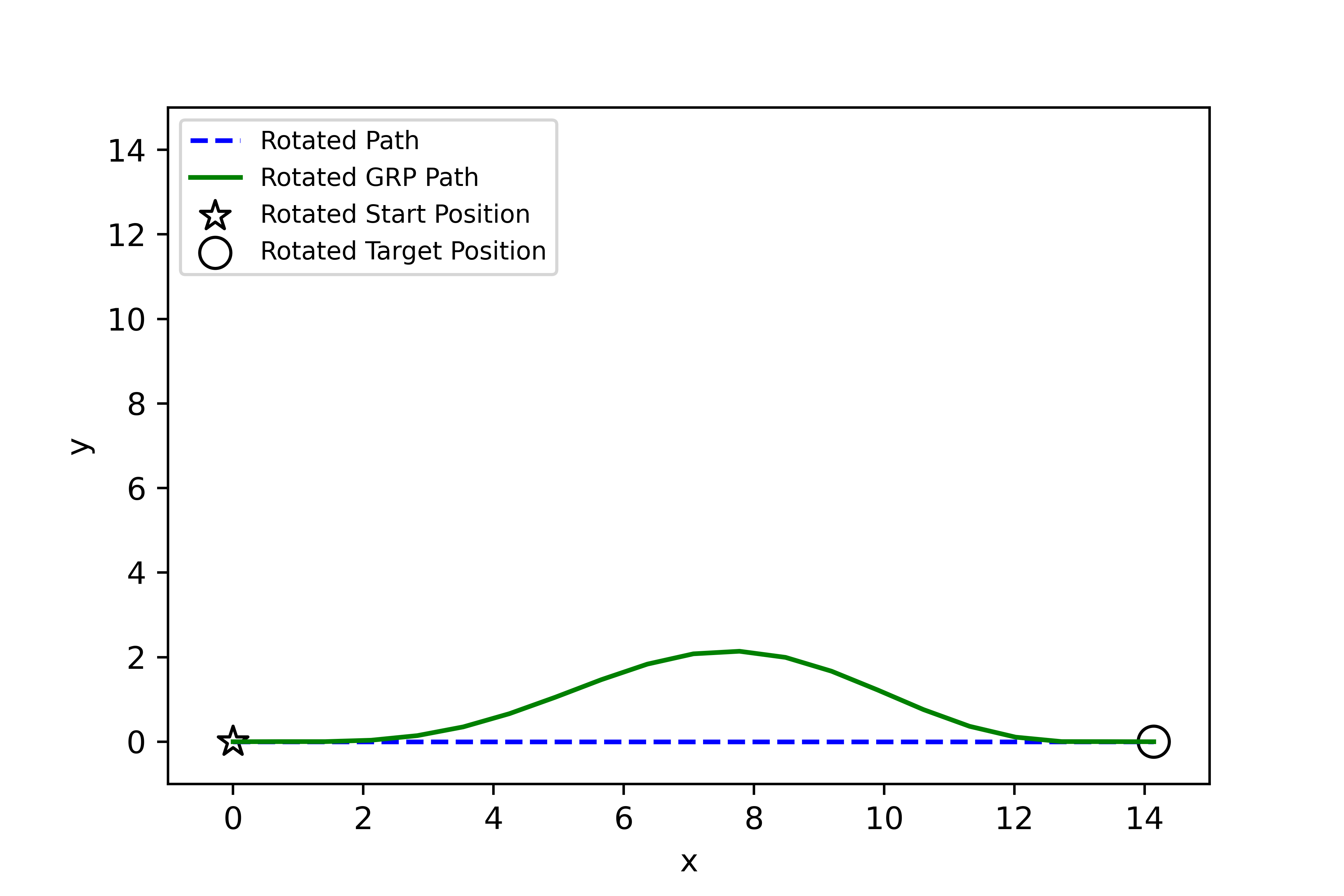}
    \caption{}
  \end{subfigure}
  \begin{subfigure}{0.475\textwidth}
    \centering
    \includegraphics[width=\textwidth]{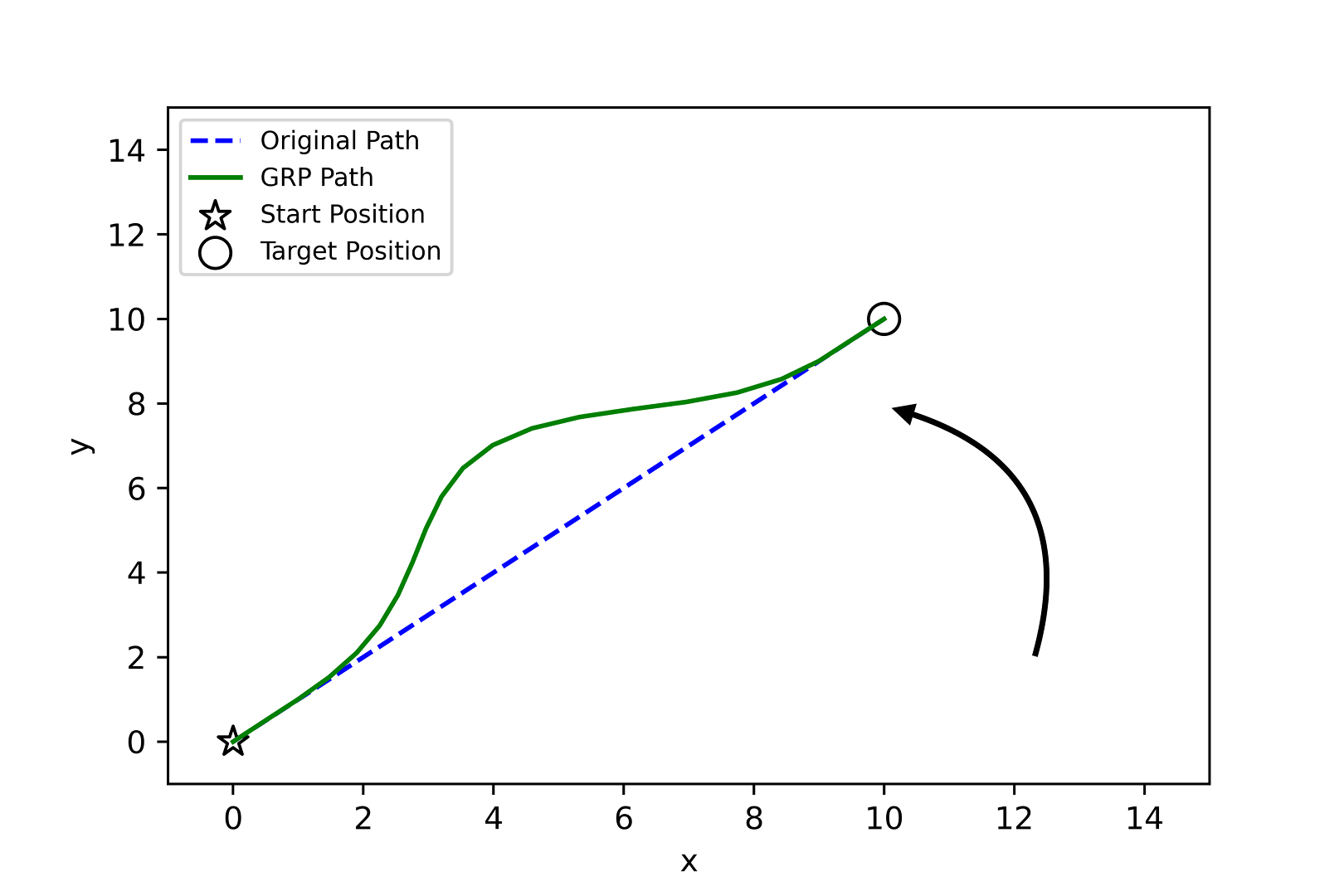}
    \caption{}
  \end{subfigure}
  \caption{The original trajectory is rotated to align the $+X$-axis (character's front direction), then GRP is employed to sample a trajectory. The generated trajectory is rotated back to match the original trajectory's starting and target position. The root joint's trajectory is visualized.}
  \label{fig:grp2d}
\end{figure}

In order to augment motion trajectory from ground truth motion, we first rotate the original trajectory to make starting and target positions aligned on the $X$-axis to facilitate the GRP process. In the rotated system, the proposed motion augmentation process samples a new trajectory, which has the same $\mathbf{x}$ values but new $\tilde{\mathbf{y}}$ values from the GRP process. We have fixed the progress on $X$ and $Z$ axes since motion should maintain its starting and target position and changing the $Z$-axis empirically yields unnatural motion sequences. Therefore, motion augmentation is applied only for `walking' and `running' motions. Through the augmentation process, the rotated path $(\mathbf{x}_{\mathcal{P}},\mathbf{y}_{\mathcal{P}},\mathbf{z}_{\mathcal{P}})$ can be augmented to multiple sampled paths $(\mathbf{x}_{\mathcal{P}}, \tilde{\mathbf{y}}_{\mathcal{P}}, \mathbf{z}_{\mathcal{P}})$. On top of the positional augmentation, we further compute the orientation difference between the original and augmented trajectories in the $XY$ plane and rotate the augmented sequence accordingly on the $XY$-plane to make the generated motion face the correct direction as the original motion. Figure \ref{fig:grp2d} illustrates the process of the proposed motion augmentation method.

\subsubsection{Semantic Embedding}

Motion representation and semantic information need to be learned together to achieve semantic-conditioned motion in-betweening. To this end, we introduce semantic embedding, which projects motion's high-level semantics into a look-up table. We prepend the semantic embedding to be the first token of the motion input matrix so that the Transformer encoder can attend to this information:

\begin{equation}
    I_{\mathrm{semantic}} = \operatorname{Concat}(\mathbf{s}, I) \in \mathbb{R}^{(T+1) \times d}
\end{equation} where $\mathbf{s}$ indicates the semantic embedding. We use the extended input motion representation before feeding the Transformer encoder.

\subsection{Loss}

There are three forms of learning losses in our model: semantic loss, position loss, and rotation loss.

First, semantic loss is a reconstruction loss. The first predicted token is a special token to represent semantics in motion. We make the model reconstruct the first token for every sequence:

\begin{equation}
    \mathcal{L}_{\mathrm{semantic}} = \frac{1}{N} \sum_{N=1}^{N} \lVert \mathbf{s} - \mathbf{\hat{s}} \rVert_1
\end{equation} where $N$ is the total number of data and $\mathbf{s}$ means semantic embedding.

Position and rotation losses are the average L1 distance between predicted sequences and their ground truth values:

\begin{align}
    \mathcal{L}_{\mathrm{position}} &= \frac{1}{N} \frac{1}{T} \sum_{N=1}^{N} \sum_{t=1}^{T} \lVert \mathbf{p}_t - \mathbf{\hat{p}}_t \rVert_1 \\
    \mathcal{L}_{\mathrm{rotation}} &= \frac{1}{N} \frac{1}{T} \sum_{N=1}^{N} \sum_{t=1}^{T} \lVert \mathbf{q}_t - \mathbf{\hat{q}}_t \rVert_1
\end{align}

Our total loss is defined as:

\begin{equation}
\mathcal{L}_{\mathrm{total}} = w_1 \mathcal{L}_{\mathrm{semantic}} + w_2 \mathcal{L}_{\mathrm{position}} + w_3 \mathcal{L}_{\mathrm{rotation}}
\end{equation}

All of our losses are scaled to be approximately equal before training begins.

%
%
\subsection{Hyperparameters}
Model is trained with AdamW \cite{loshchilov2017decoupled} optimizer with a learning rate of 0.0001, $\beta_1 = 0.9$, $\beta_2 = 0.99$, and weight decay of 0.01. Also, we set Transformer encoder to have 7 heads, 8 encoder layers, 2,048 feedforward dimensions, and 0.05 dropout probability during training. Weights for the total losses $w_1, w_2$, and $w_3$ are set to 1.5, 0.05, and 2.0 respectively. In all our experiments, minibatch size is set to $32$.

%
%
\section{Experiments}
\label{sec:experiment_and_results}

This section demonstrates the experimental settings and presents qualitative and quantitative results. Please refer to our project web page\footnote{\url{https://jihoonerd.github.io/Conditional-Motion-In-Betweening}} for more videos from the experiments. 

%
%
\subsection{Datasets}
We evaluate our algorithm on four widely-used motion datasets for motion in-betweening (MIB) task: LAFAN1 \cite{harvey2020robust}, HumanEva \cite{sigal2010humaneva}, HUMAN4D \cite{chatzitofis2020human4d}, and MPI-HDM05 \cite{cg-2007-2}. We split the training and test datasets by subjects which is described in Table \ref{table:train-test-split}.

\textbf{LAFAN1 \cite{harvey2020robust}:} LAFAN1 is a high-quality public motion capture dataset. It contains $15$ actions such as walking, dancing, fighting, jumping, etc, with approximately $4.6$ hours long and contains $496,672$ frames taken by five actors. It was used in Harvey et al. \cite{harvey2020robust}, which first proposed a robust MIB method. We evaluate the performance of our proposed method using this dataset for both unconstrained MIB and CMIB tasks.

\textbf{HumanEva \cite{sigal2010humaneva}:} This dataset is obtained from a 3D whole-body motion capture system. It is recorded by four subjects with six common actions such as walking, jogging, gesturing, etc. Three RGB cameras and four grayscale cameras are used for the recording.

\textbf{HUMAN4D \cite{chatzitofis2020human4d}:} This is a recently-collected large and multimodal public motion dataset with various human actions. This dataset is captured from $24$ MoCap cameras with four depth sensors. The full dataset is composed of $50,306$ frames with $19$ activity labels.

\textbf{MPI-HDM05 \cite{cg-2007-2}:} MPI-HDM05 is a research-purpose dataset especially for motion analysis, synthesis, and classification. It contains more than three hours of recorded motion frames.

\begin{table}[h]
\caption {Experiment configuration of train and test data.} \label{table:train-test-split}
\centering
\begin{tabular}{c|c|c}
    Dataset & Train (subject) & Test (subject) \\ \hline
    \textbf{LAFAN1} & 1, 2, 3, 4 & 5 \\ \hline
    \textbf{HumanEva} & 1, 2 & 3 \\ \hline
    \textbf{HUMAN4D} & 1, 2, 3, 4, 5, 6, 7 & 8 \\ \hline
    \textbf{MPI-HDM05} & bk, dg, mm & tr \\ \hline
\end{tabular}
\end{table}

%
%
\subsection{Motion Data Pre-processing}
We adapt the same pre-processing procedures with \cite{harvey2020robust} for a fair comparison. In particular, we split each motion into a fixed-width window where all windowed motion sub-sequences are rotated to align $X^{+}$ axis ahead at the $10$-th frame. For semantic conditioning, we choose labeled actions as semantic information of the motion. Positions and rotations are represented in the global coordinate frame.
We also build a deploy pipeline for the Unity Engine\footnote{\url{https://unity.com}} to present qualitative results.

%
%
\subsection{Motion In-betweening Results}

We use L2P and L2Q used in \cite{harvey2020robust} to evaluate our model's MIB performance. L2P and L2Q are the average L2 norm of global positions and quaternions:
\begin{align}
    \mathrm{L2P} = \frac{1}{N} \frac{1}{T} \sum_{N=1}^{N} \sum_{t=1}^{T} \lVert \mathbf{p}_{t} - \hat{\mathbf{p}}_{t} \rVert_2 \\ 
    \mathrm{L2Q} = \frac{1}{N} \frac{1}{T} \sum_{N=1}^{N} \sum_{t=1}^{T} \lVert \mathbf{q}_{t} - \hat{\mathbf{q}}_{t} \rVert_2
\end{align}
where $N$ indicates the number of data being evaluated. Note that global positions are normalized before calculating L2P.

Here, we compare our model with the state-of-the-art robust motion in-betweening models in MIB tasks. Since the implementation of RMIB \cite{harvey2020robust} and SSMCT \cite{duan2021single} are not publicly available, the quantitative results of both models that are not reported in the original paper are provided by our own reproduced implementation. We would like to note that our reproduced implementation also shows similar results in their reported performance. 

\begin{table}[h]
\caption{Motion in-betweening results on LAFAN1.}
\label{table:in-betweening}
\begin{subtable}[h]{\textwidth}
\centering
\begin{tabular}{c|c|c|c|c|c}
    Length (frames) & 30 & 40 & 60 & 80 & 120\\ \hline \hline
    Zero-Velocity & 6.60 & 8.44 & 10.20 & 13.42 & 18.69\\ \hline
    Interp. & 2.32 & 3.05 & 3.87 & 5.63 & 9.12 \\ \hline
    Bi-LSTM & 1.56 & 2.13 & 3.54 & 4.80 & 8.91 \\ \hline
    ERD-QV \cite{harvey2020robust} & 1.28 & 2.03 & 3.28 & 4.51 & 7.61 \\ \hline
    CAE \cite{kaufmann2020convolutional} & 1.23 & 1.64 & 2.68 & 3.46 & 5.43 \\ \hline
    SSMCT \cite{duan2021single} & \textbf{1.10} & \textbf{1.33} & 2.04 & 2.78 & 4.32 \\ \hline
    CMIB (Ours) & 1.19 & 1.35 & \textbf{1.98} & \textbf{2.70} & \textbf{4.25} \\ \hline
\end{tabular}
\caption{L2P}
\end{subtable}
\begin{subtable}[h]{\textwidth}
\centering
\begin{tabular}{c|c|c|c|c|c}
Length (frames) & 30 & 40 & 60 & 80 & 120\\ \hline \hline
Zero-Velocity & 1.51 & 1.72 & 1.91 & 2.22 & 2.67 \\ \hline
Interp. & 0.98 & 1.15 & 1.34 & 1.72 & 2.28 \\ \hline
Bi-LSTM & 0.79 & 0.88 & 1.25 & 1.52 & 2.07 \\ \hline
ERD-QV \cite{harvey2020robust} & 0.69 & 0.79 & 1.12 & 1.48 & \textbf{1.51} \\ \hline
CAE \cite{kaufmann2020convolutional} & 0.84 & 0.95 & 1.30 & 1.65 & 2.10 \\ \hline
SSMCT \cite{duan2021single} & 0.61 & 0.76 & 1.14 & 1.53 & 1.98 \\ \hline
CMIB (Ours) & \textbf{0.59} & \textbf{0.71} & \textbf{1.04} & \textbf{1.42} & 1.90 \\ \hline
\end{tabular}
\caption{L2Q}
\end{subtable}
\end{table}

\begin{figure}[h]
    \centering
    \includegraphics[width=0.475\textwidth]{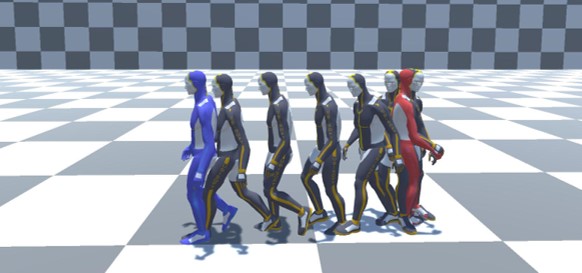}
    \includegraphics[width=0.475\textwidth]{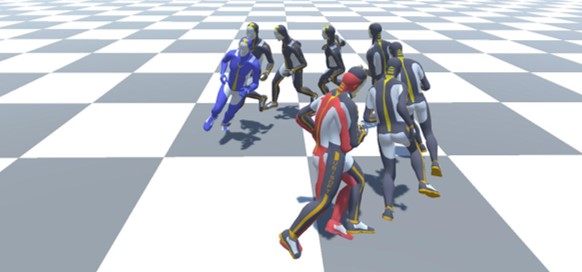}
    \caption{Unconstrained in-betweening results. Intermediate poses are generated from starting pose (red) to target pose (blue).}
    \label{fig:in-betweening}
\end{figure}

Our model is compared with zero-velocity, interpolation, bidirectional LSTM model, ERD-QV \cite{harvey2020robust}, CAE \cite{kaufmann2020convolutional}, and SSMCT \cite{duan2021single}. The zero-velocity baseline interpolates the missing frames with the latest frame. The interpolation baseline model is the same as our baseline pose interpolation, which uses LERP for joints' positional values and SLERP for rotational (quaternion) values. Table \ref{table:in-betweening} shows the MIB evaluation results on the test dataset of LAFAN1, and Table \ref{table:other-datasets} presents L2P and L2Q score comparison on other public motion datasets. Our proposed method presents comparable or better performances in most MIB tasks, even with the additional controllability. Synthesized motions are visualized in Figure \ref{fig:in-betweening}.

\begin{table}[h]
\caption{Motion in-betweening results on other motion datasets.} \label{table:other-datasets}
\centering
\begin{tabular}{c|c|c|c|c|c|c|c}
\multicolumn{2}{c|}{Length (frames)} & \multicolumn{2}{c|}{40} & \multicolumn{2}{c|}{60} & \multicolumn{2}{c}{80} \\ \hline
\multicolumn{2}{c|}{Metric} & L2P & L2Q & L2P & L2Q & L2P & L2Q \\ \hline \hline
\multirow{5}{*}{HumanEva \cite{sigal2010humaneva}} & Bi-LSTM & 3.71 & 0.80 & 4.51 & 1.30 & 4.92 & 1.57 \\ \cline{2-8}
& ERD-QV \cite{harvey2020robust} & 3.58 & 0.73 & 4.32 & 1.26 & 4.73 & 1.42 \\ \cline{2-8}
& CAE \cite{kaufmann2020convolutional} & 3.49 & 0.83 & 3.98 & 1.42 & 4.65 & 1.73 \\ \cline{2-8}
& SSMCT \cite{duan2021single} & \textbf{3.42} & 0.73 & 3.88 & 0.85 & 4.52 & 0.96 \\ \cline{2-8}
& CMIB (Ours) & 3.45 & \textbf{0.70} & \textbf{3.84} & \textbf{0.80} & \textbf{4.48} & \textbf{0.87} \\ \hline
\multirow{5}{*}{HUMAN4D \cite{chatzitofis2020human4d}} & Bi-LSTM & 3.48 & 0.83 & 5.11 & 1.08 & 6.23 & 1.24 \\ \cline{2-8}
& ERD-QV \cite{harvey2020robust} & 3.40 & 0.79 & 4.95 & 0.99 & 5.93 & \textbf{1.15} \\ \cline{2-8}
& CAE \cite{kaufmann2020convolutional} & 3.29 & 0.87 & 4.78 & 1.13 & 5.34 & 1.35 \\ \cline{2-8}
& SSMCT \cite{duan2021single} & 3.11 & 0.71 & 4.10 & 1.02 & 4.80 & 1.27 \\ \cline{2-8}
& CMIB (Ours) & \textbf{3.10} & \textbf{0.70} & \textbf{4.02} & \textbf{0.95} & \textbf{4.66} & 1.17 \\ \hline
\multirow{5}{*}{MPI-HDM05 \cite{cg-2007-2}} & Bi-LSTM & 2.65 & 0.86 & 3.80 & 1.23 & 4.72 & 1.40 \\ \cline{2-8}
& ERD-QV \cite{harvey2020robust} & 2.60 & 0.82 & 3.63 & \textbf{1.08} & 4.60 & \textbf{1.26} \\ \cline{2-8}
& CAE \cite{kaufmann2020convolutional} & 2.54 & 0.92 & 3.58 & 1.39 & 4.41 & 1.52 \\ \cline{2-8}
& SSMCT \cite{duan2021single} & 2.33 & 0.85 & 3.18 & 1.21 & 3.85 & 1.49 \\ \cline{2-8}
& CMIB (Ours) & \textbf{2.30} & \textbf{0.81} & \textbf{3.02} & 1.15 & \textbf{3.78} & 1.45 \\ \hline
\end{tabular}
\end{table}

\begin{figure}[h]
     \centering
     \begin{subfigure}[b]{0.475\textwidth}
         \centering
         \includegraphics[width=\textwidth]{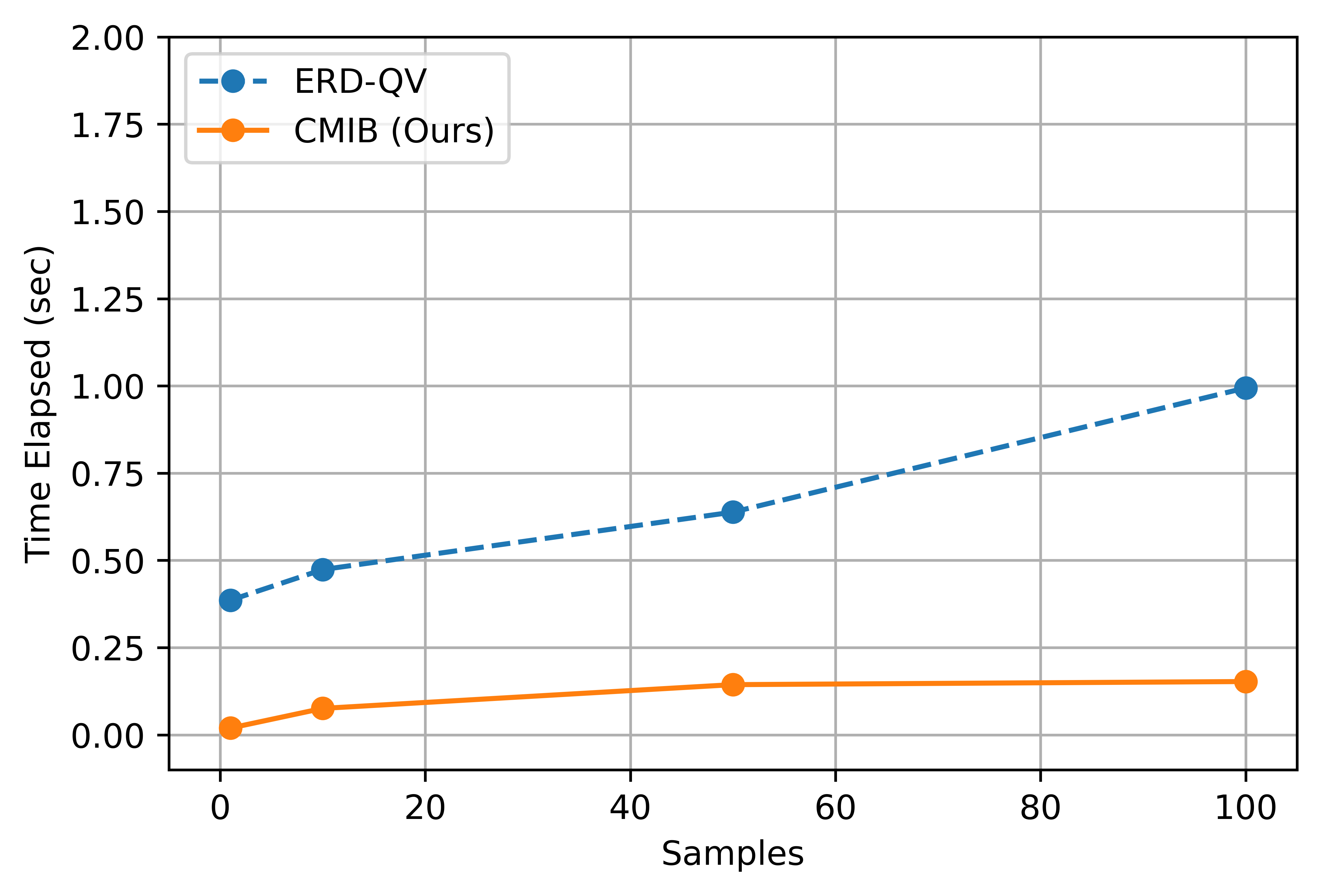}
         \caption{In-betweening length: 30 frames}
     \end{subfigure}
     \hfill
     \begin{subfigure}[b]{0.475\textwidth}
         \centering
         \includegraphics[width=\textwidth]{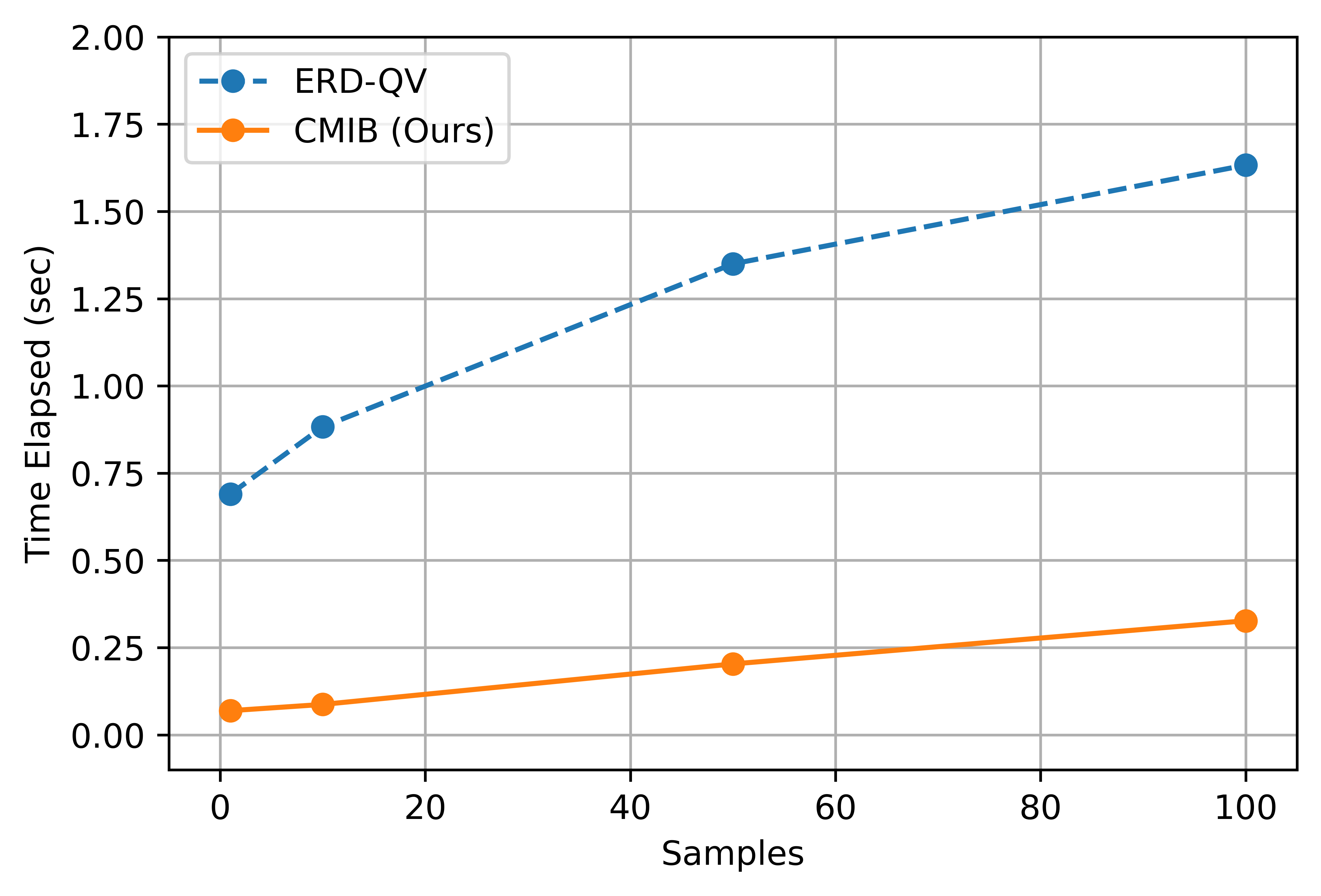}
         \caption{In-betweening length: 80 frames}
     \end{subfigure}
     \hfill
    \caption{Speed performance. CPU inference time is measured with Intel\textsuperscript{\tiny\textregistered} Xeon\textsuperscript{\tiny\textregistered} Gold 5120 CPU @ 2.20GHz. Inference time is measured as an average after 30 trials.}
    \label{fig:speed-comparison}
\end{figure}

Figure \ref{fig:speed-comparison} presents the computation time of different batch sizes and in-betweening frames. This supports that the non-autoregressive model has a great advantage in inference time speed performance over the autoregressive models.

%
%
\subsection{Pose-conditioned Motion In-betweening Results}

Pose-conditioned MIB is a process of generating natural skeletal motion with the start, target, and additional anchor poses constraints which can be thought of as having additional anchor pose constraints to the unconstrained MIB task. Since a non-autoregressive Transformer architecture of utilized, we can leverage multiple numbers of anchor poses while training. In this paper, however, we simply assume that a single anchor pose is given for the pose-conditioned MIB task, and while training, an anchor pose is uniformly sampled within the horizon of a motion sequence (i.e., $t \sim \mathcal{U}(1, T)$ where $T$ is the sequence length). 

\begin{figure}[h]
\centering
    \begin{subfigure}{0.875\textwidth}
        \centering
        \includegraphics[width=\textwidth]{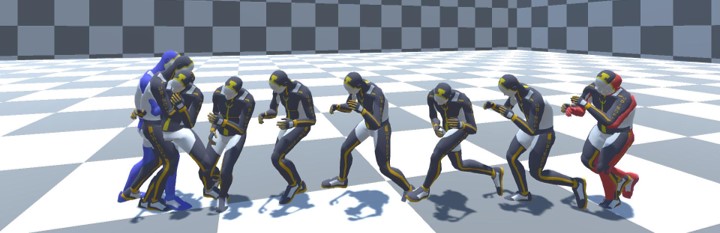}
    \end{subfigure}
    \begin{subfigure}{0.875\textwidth}
        \centering
        \includegraphics[width=\textwidth]{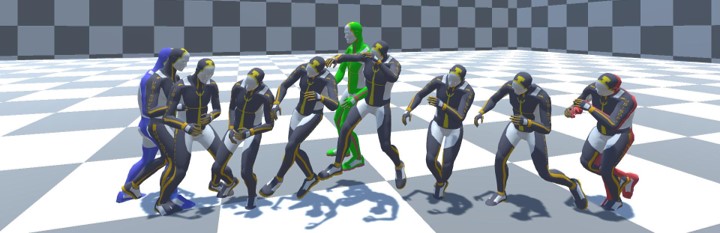}
    \end{subfigure}
    \caption{(Top) Unconstrained MIB result from the start pose (red) to the target pose (blue). (Bottom) Pose-conditioned MIB result with an additional anchor pose (green)}
    \label{fig:pc}
\end{figure}

Note that the pose-conditioned MIB results are subjected to the diversity of trajectories within the dataset, where we observe that pose-conditioned generation often fails when the anchor pose is located outside the plausible reason. However, this limitation can be alleviated via the proposed motion data augmentation method in Section \ref{subsec:Data_Aug}. To validate the effectiveness of the proposed motion data augmentation on the pose-conditioned MIB tasks, we select the poses at $t=20, \, 40, \, \text{and} \, 60$ on aiming, dance, fight, run, and walk motions of LAFAN1 (total $670$ motions). Then, we add random perturbations to the $(x,\,y)$ position of the anchor pose and check how the generated sequences pass through the given anchor pose in the $XY$ plane. We further change the orientation of the start and target poses to match the given anchor pose. 

Table \ref{table:pose-conditioning} shows the comparative results of the pose-conditioned MIB task with and without the proposed motion augmentation method in terms of the L2 norm between the given and generated root positions at $t=20, \, 40, \, \text{and} ~ 60$. Our proposed augmentation method greatly improves the pose-conditioned MIB performance on all anchor positions. Figure \ref{fig:pc} depicts the pose-conditioned MIB result. When given a jumping anchor pose (green), the entire generated motion changes accordingly to make the jump motion. 

\begin{table}[h]
\caption {Ablation study on augmentation. L2 distance at anchor pose.} \label{table:pose-conditioning}
\centering
\begin{tabular}{c|c|c|c}
    Anchor Position & \multicolumn{1}{c|}{$t=20$} & \multicolumn{1}{|c|}{$t=40$} & \multicolumn{1}{|c}{$t=60$} \\ \hline
    CMIB (w/o Aug) & 0.35 & 0.78 & 0.35 \\ \hline
    CMIB (w/ Aug) & 0.18 & 0.18 & 0.19 \\ \hline
\end{tabular}
\end{table}

%
%
\subsection{Semantic-conditioned Motion In-betweening Results}

\begin{figure}[h]
     \centering
     \begin{subfigure}[h]{0.8\textwidth}
         \centering
         \includegraphics[width=0.8\textwidth]{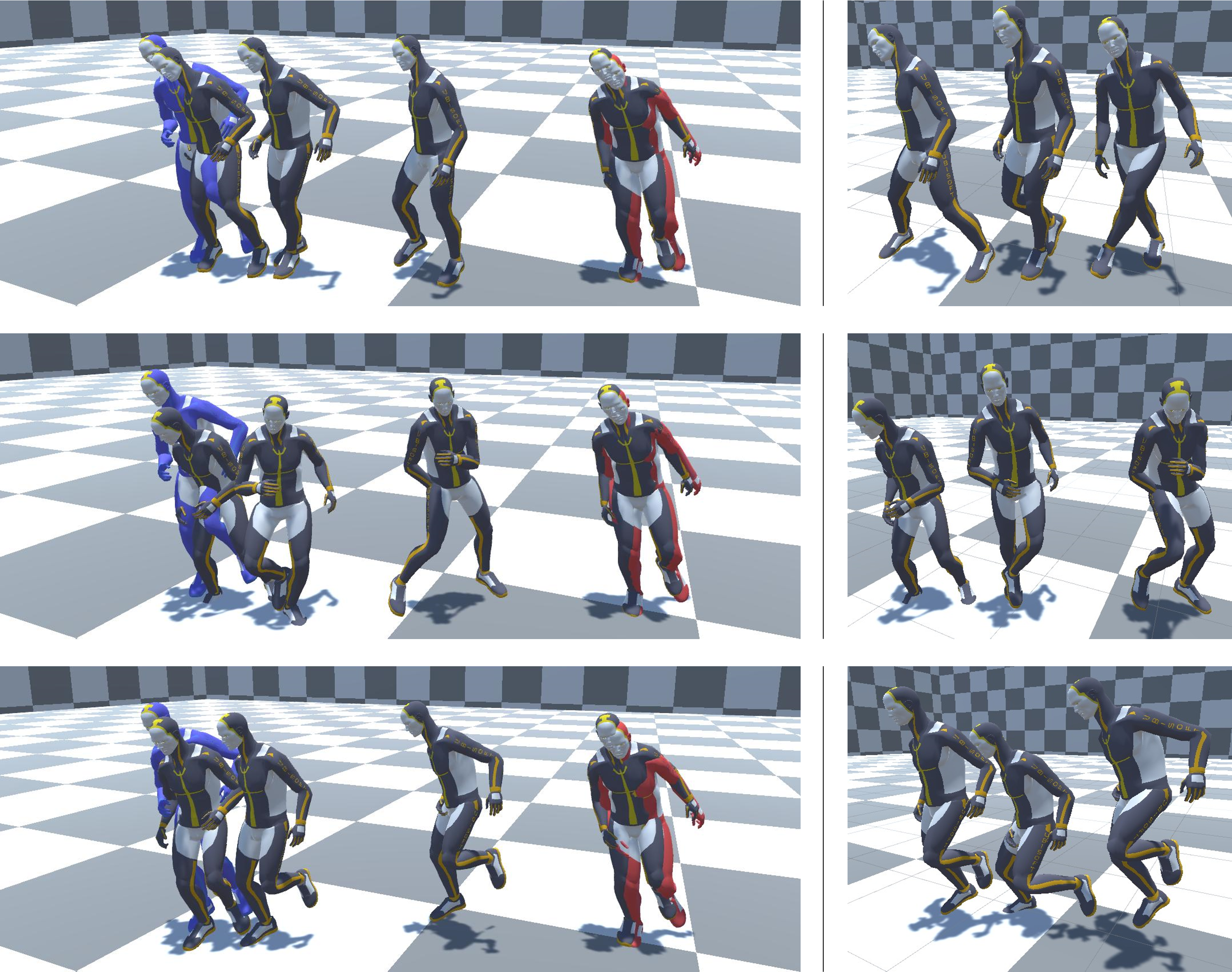}
         \caption{}
         \label{fig:sc-sample-1}
     \end{subfigure}
     \hfill
     \begin{subfigure}[h]{0.8\textwidth}
         \centering
         \includegraphics[width=0.8\textwidth]{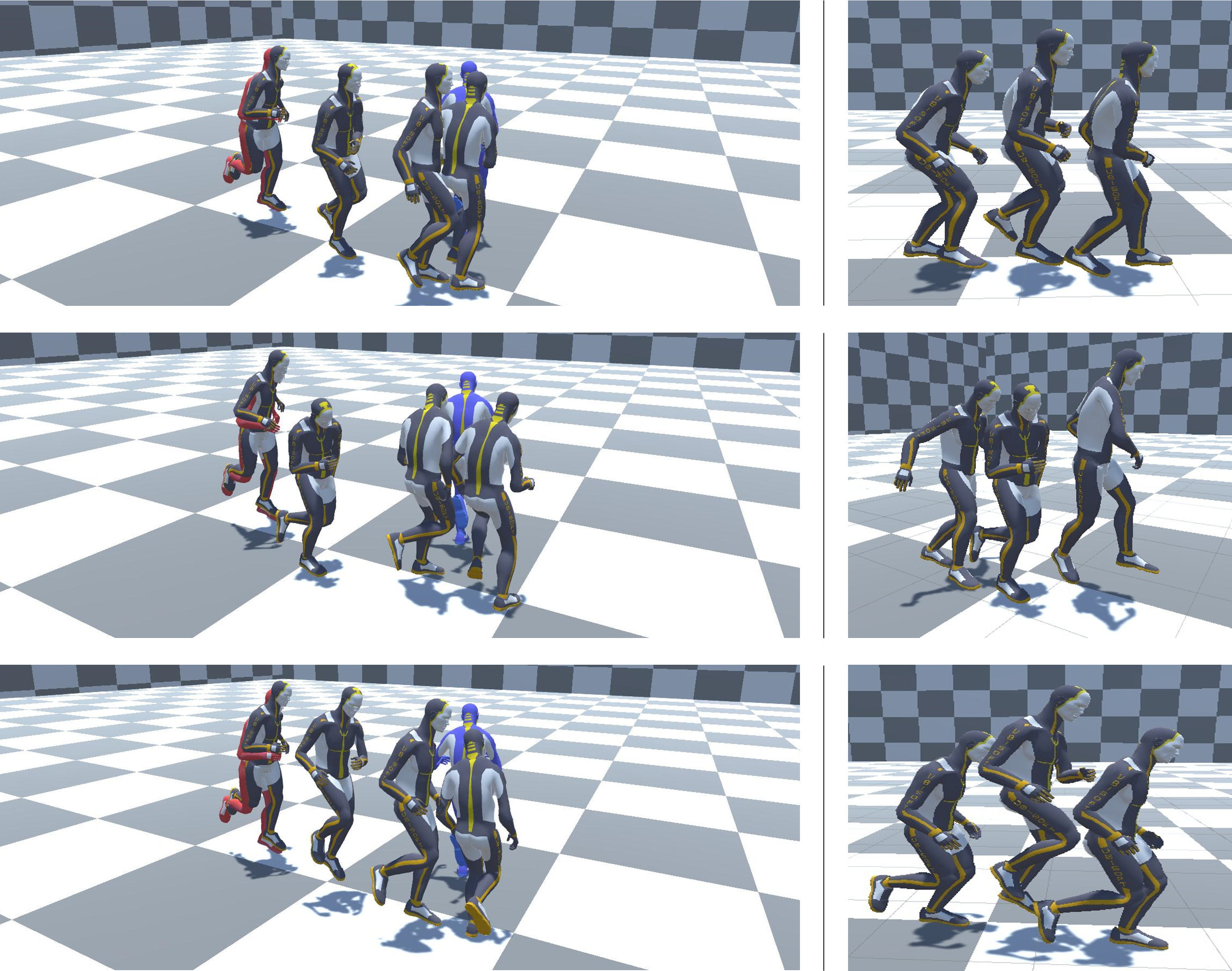}
         \caption{}
         \label{fig:sc-sample-1}
     \end{subfigure}
     \hfill
    \caption{Semantic-conditioned in-betweening results. Conditioning semantics are walk (top), dance (middle), and jump (bottom). The left column depicts the entire sequence trajectory with starting (red) and target (blue) poses. The right column visualizes interim three consecutive poses for visual comparison between different semantics.}
    \label{fig:se-conditioned}
\end{figure}

In this experiment, we set conditioning semantics as $15$ action labels provided from the LAFAN1 dataset. We evaluate our model based on the reconstruction error of the semantic token values. For quantitative evaluation, we divide the test dataset by action labels and make the model conduct MIB for available action labels. Pose conditioning is disabled for the experiments (i.e., no anchor pose is given). Here we evaluate on the model's performance in terms of L2P. We assume that the corresponding ground truth labels should yield the smallest error. Table \ref{table:semantic-l2p} describes the results and Figure \ref{fig:se-conditioned} visualizes the results of semantic-conditioned MIB.

From the visualized results, we found that infilling horizon affects the visual quality of semantic-conditioned MIB. Long-horizon MIB tends to produce better results compared to the short-horizon task in the semantic-conditioned generation. For example, in the case of jumping motions, it is difficult to generate a complete jumping motion with a 30-frames horizon, but long-horizon infilling is more likely to produce clear jumping motions. This suggests that there will be a minimum number of frames to reflect the semantics of motion.

\setlength{\tabcolsep}{2.0pt}
\begin{table}[h]

\caption {Semantic-conditioned motion in-betweening results. Column-wise direction indicates conditioning semantics provided to model and row-wise direction means ground truth conditions in the test dataset.} \label{table:semantic-l2p}

\begin{subtable}{0.45\textwidth}
\begin{tabular}{c|ccccc}
    & walk & run & dance & jump & fight \\ \hline
    walk & 1.03 & \textbf{1.02} & 1.26 & 1.15 & 1.11 \\ \hline
    run & 1.35 & \textbf{1.31} & 1.58 & 1.60 & 1.37 \\ \hline
    dance & 1.64 & \textbf{1.63} & 1.69 & 1.66 & 1.64 \\ \hline
    jump & 1.64 & 1.67 & 1.95 & \textbf{1.34} & 1.73 \\ \hline
    fight & 1.82 & \textbf{1.81} & 2.11 & 1.89 & 1.82 \\ \hline
\end{tabular}
\caption{L2P: 40 Frames}
\end{subtable}
\hfill
\begin{subtable}{0.45\textwidth}
\begin{tabular}{c|ccccc}
    & walk & run & dance & jump & fight \\ \hline
    walk & \textbf{1.65} & 1.77 & 1.98 & 1.87 & 1.77 \\ \hline
    run & 2.00 & \textbf{1.92} & 2.29 & 2.21 & 2.00 \\ \hline
    dance & 2.02 & 2.04 & 2.14 & \textbf{2.01} & 2.08 \\ \hline
    jump & 2.12 & 2.20 & 2.38 & \textbf{1.83} & 2.22 \\ \hline
    fight & 2.51 & \textbf{2.49} & 2.84 & 2.58 & 2.50 \\ \hline
\end{tabular}
\caption{L2P: 60 Frames}
\end{subtable}
\hfill
\begin{subtable}{0.45\textwidth}
\begin{tabular}{c|ccccc}
    & walk & run & dance & jump & fight \\ \hline
    walk & \textbf{2.43} & 2.50 & 2.74 & 2.60 & 2.52 \\ \hline
    run & 3.13 & \textbf{2.92} & 3.40 & 3.29 & 2.97 \\ \hline
    dance & 2.42 & 2.43 & 2.52 & \textbf{2.37} & 2.46 \\ \hline
    jump & 2.60 & 2.82 & 2.87 & \textbf{2.38} & 2.78 \\ \hline
    fight & 3.40 & 3.44 & 3.67 & 3.50 & \textbf{3.34} \\ \hline
\end{tabular}
\caption{L2P: 80 Frames}
\end{subtable}
\hfill
\begin{subtable}{0.45\textwidth}
\begin{tabular}{c|ccccc}
    & walk & run & dance & jump & fight \\ \hline
    walk & \textbf{4.43} & 4.45 & 4.76 & 4.79 & 4.81 \\ \hline
    run & 5.98 & \textbf{5.58} & 6.40 & 6.55 & 6.27 \\ \hline
    dance & \textbf{3.68} & 3.89 & 3.76 & 3.69 & 3.93 \\ \hline
    jump & 3.63 & 4.59 & 3.84 & \textbf{3.43} & 3.86 \\ \hline
    fight & 5.02 & 5.10 & 5.21 & 5.14 & \textbf{4.86} \\ \hline
\end{tabular}
\caption{L2P: 120 Frames}
\end{subtable}
\end{table}

%
%
\section{Limitation and Discussion}
\label{sec:limitation-and-discussion}

\begin{figure}[h]
\centering
\includegraphics[width=0.875\textwidth]{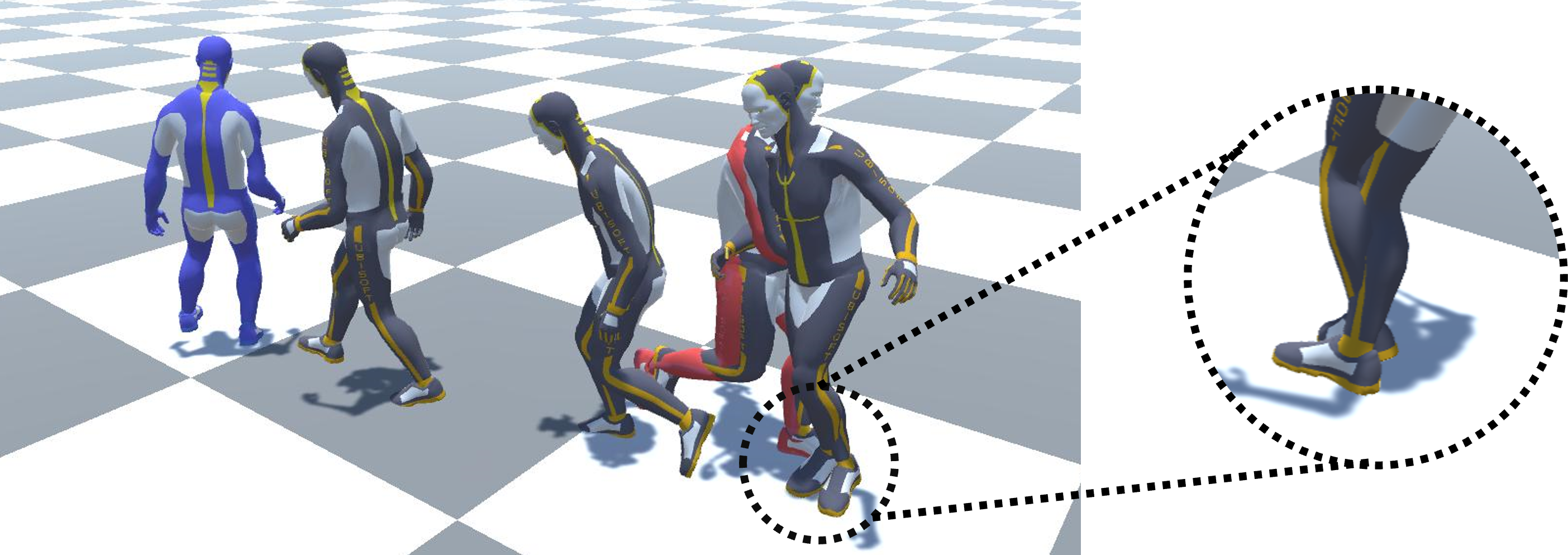}
\caption{Motion in-betweening results with a penetration.}
\label{fig:limitation-penetration}
\end{figure}

\begin{figure}[h]
\centering
\includegraphics[width=0.875\textwidth]{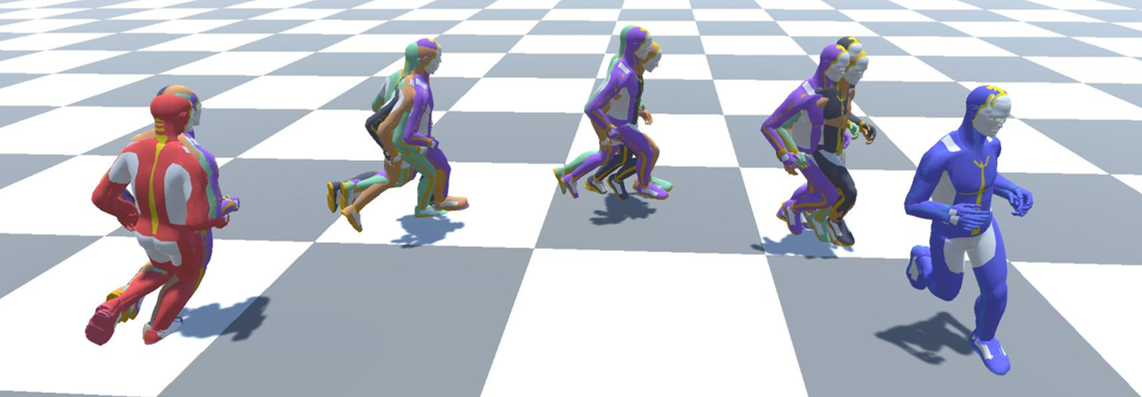}
\caption{Failure case for semantic control. Grey, orange, green, and purple are conditionally generated from `walk', `jump', `fight', and `dance' labels, respectively. Red and blue poses are starting and target poses.}
\label{fig:limitation-conditioning-failure}
\end{figure}

In this work, we confirm that the CMIB model can perform the most of in-betweening settings, but there are some limitations. First, generated motion does not take volume into consideration, which produces motion with penetration between the human body as visualized in Figure \ref{fig:limitation-penetration}. It can be mitigated by including a post-processing pipeline or integrating loss term for penetrating configuration as a regularizer. The other failure case is conditioning failure. This case is similar to mode collapse in GAN, synthesizing nearly identical motion regardless of conditioning semantics. We suspect that the problem comes from two objectives that are difficult to satisfy simultaneously. For instance, if given starting and target poses are too far apart, motion in-betweening objective mandates a model to generate running motion without regard to given semantics such as `fight' or `dance'. This is depicted in Figure \ref{fig:limitation-conditioning-failure}. Empirically, conditioned generation performs best when given starting and target poses are feasible enough to generate the given condition.

%
%
\section{Conclusion}
\label{sec:conclusion}

In this study, we have focused on the problem of adding controllability to motion in-betweening (MIB) tasks. In particular, we proposed a conditional motion in-betweening (CMIB) method using non-autoregressive Transformer encoder architecture where pose-conditioned and semantic-conditioned MIB methods are achieved via randomized shuffled anchor pose and semantic embedding token, respectively. An effective motion data augmentation method is also presented to increase the performance of the pose-conditioned MIB. Our proposed method outperforms existing state-of-the-art MIB with respect to pose prediction accuracy and inference time while providing additional controllability.

While our approach demonstrates controllability in MIB, there are still promising areas where more research is required. As our model relies on the non-autoregressive structure, the horizon it predicts is limited to the predefined maximum length, and the evaluation metric is not suitable for expressing the diversity of human motions. It would be interesting to combine our method with explicit density estimation methods to better capture the possible human motion distributions. Furthermore, we will continue to extend controllable motion generation combined with other modalities such as natural language commands.

\section*{Acknowledgements}
This work was supported by Institute of Information \& communications Technology Planning \& Evaluation (IITP) grant funded by the Korea government(MSIT) (No. 2019-0-00079 , Artificial Intelligence Graduate School Program(Korea University)).


\bibliographystyle{elsarticle-num}
\bibliography{cas-refs}





\end{document}